\documentclass[12pt,letterpaper]{article}
\usepackage[a4paper, total={7in, 10in}]{geometry}

\usepackage{graphicx}
\usepackage{helvet}
\usepackage{authblk}
\usepackage{hyperref}
\usepackage{amsmath} 
\usepackage{amssymb} 
\usepackage{comment}
\usepackage{cleveref}
\usepackage{algorithm}
\usepackage{subcaption}
\usepackage{caption}
\usepackage[super,comma,sort&compress]  
   {natbib}
\usepackage[right]{lineno} \linenumbers

\makeatletter
\renewcommand{\maketitle}{\bgroup\setlength{\parindent}{0pt}
\begin{flushleft}
  \textbf{\@title}
  
  \@author
\end{flushleft}\egroup}
\makeatother

\title{Fine-Pruning: A Biologically Inspired Algorithm for Personalization of Machine Learning Models}
\date{}

\author[1]{Joseph Bingham}
\author[2]{Saman Zonouz}
\author[1,3]{Dvir Aran}

\affil[1,*]{Faculty of Biology, Technion - Israel Institute of Technology, Haifa, IL}
\affil[2]{College of Computing, The Georgia Institute of Technology, Atlanta, GA, USA}
\affil[3]{Taub Faculty of Computer Science, Technion - Israel Institute of Technology, Haifa, IL}

\affil[*]{Correspondence: jbingham@campus.technion.ac.il}

\begin{document}

\maketitle

\section*{SUMMARY}
  
Neural networks have long strived to emulate the learning capabilities of the human brain. While deep neural networks (DNNs) draw inspiration from the brain in neuron design, their training methods diverge from biological foundations. Backpropagation, the primary training method for DNNs, requires substantial computational resources and fully labeled datasets, presenting major bottlenecks in development and application. This work demonstrates that by returning to biomimicry, specifically mimicking how the brain learns through pruning, we can solve various classical machine learning problems while utilizing orders of magnitude fewer computational resources and no labels. Our experiments successfully personalized multiple speech recognition and image classification models, including ResNet50 on ImageNet, resulting in increased sparsity of approximately 70\% while simultaneously improving model accuracy to around 90\%, all without the limitations of backpropagation. This biologically inspired approach offers a promising avenue for efficient, personalized machine learning models in resource-constrained environments.

\section*{KEYWORDS}


Machine Learning, Biomimicry, Neurosynaptic Pruning, Biologically Feasible Learning

\section*{Introduction}

The pursuit of artificial intelligence (AI) that mimics human cognitive abilities has long been a driving force in computer science. At the heart of this quest lies the development of artificial neural networks, computational models inspired by the intricate web of neurons in the human brain. These artificial neural networks, particularly in their deep and complex forms known as Deep Neural Networks (DNNs)~\cite{lecun2015deep,wang2023scientific}, have revolutionized fields ranging from healthcare~\cite{CHEN20181241} and finance~\cite{sezer2020financial} to agriculture~\cite{kamilaris2018deep} and speech recognition~\cite{garg2020hierarchical}. The impact of DNNs is perhaps most tangibly felt in the realm of consumer technology, with applications like automatic speech recognition becoming ubiquitous since the debut of SIRI~\cite{hoy2018alexa} on smartphones in 2011~\cite{yu2016automatic,lee1988automatic}. Despite their biological inspiration, the learning mechanisms employed by DNNs have diverged significantly from their neurological counterparts. The primary method for training these networks, known as backpropagation~\cite{werbos1990backpropagation}, relies heavily on large amounts of labeled data and substantial computational resources~\cite{lillicrap2020backpropagation,BALDI20181,backpropagation}. This approach, while effective, presents significant challenges in real-world applications, particularly when deploying models on resource-constrained devices or in scenarios where labeled data is scarce, which often requires greater domain knowledge to be used as part of the training process to compensate~\cite{bingham2022guide}. Moreover, the generalized nature of these models often fails to capture the nuanced variations present in individual users' data, leading to suboptimal performance in personalized applications~\cite{sim2019investigation}.

Recent trends in machine learning have seen a renewed interest in biologically feasible learning algorithms~\cite{10.7554/eLife.20899,DBLP:journals/corr/abs-1804-08150,taherkhani2020review}, seeking to bridge the gap between artificial and biological neural networks. This resurgence is driven by the remarkable efficiency and adaptability of the human brain, which continues to outperform artificial systems in many aspects of cognition. One particularly promising avenue of research draws inspiration from the process of neural pruning observed in the developing human brain. Synaptic pruning~\cite{chechik1999neuronal}, a fascinating aspect of brain development, occurs during critical periods of neural development, particularly in early childhood and adolescence. This process, counter-intuitively, involves the elimination of neural connections, with the brain losing up to 50\% - 70\% of its synapses in some regions~\cite{pmid20972254,Sekar2016-hp}. However, synaptic pruning is not a random process of neural attrition, but rather a refined mechanism of neural optimization. It preferentially eliminates weak or rarely used connections while preserving and strengthening frequently activated pathways~\cite{pmid25742003}. This biological 'use it or lose it' principle results in more efficient neural circuits, improved signal-to-noise ratios, and enhanced cognitive abilities~\cite{Sakai2020-wm}. The brain's ability to sculpt its neural architecture in response to experience and specific environmental demands provides a powerful model for adaptive learning systems. During early stages of development, the brain systematically eliminates unused or weak neural connections, a process that paradoxically enhances cognitive function by reducing noise and focusing neural activity on the most relevant pathways~\cite{synaps_pruning}. This biological phenomenon provides an intriguing approach to addressing the challenges of model personalization and resource efficiency in deep neural networks.

The impetus for searching for an alternative method for personalizing a DNN comes from the observations that the running of conventional backpropagation requires orders of magnitude more memory than even advanced Internet of Thing (IoT) devices have available. As an example, an Apple iPhone has less than 2.5 billion of FLOPs~\cite{apple1}, whereas a model such as VGG-19 requires more than 20 billion FLOPs to train. Even modest architecture like ResNet-50 requires around 4 GFLOPs, twice as many as are available~\cite{ijcai2018p336}. Even if a user would accept the time required to train such models, the energy consumption would still be a major bottleneck for their deployment on such devices. Looking again at power resources consumed by a typical application~\cite{reddit}, training MobileNet, which was designed to be used on mobile devices, takes over 2x the amount of power~\cite{10.3389/fnins.2016.00496}. Looking at a the power consumption of a more conventional network, such as DenseNet, it requires 10x as much power as a typical application, which is over 20\%~\cite{apple1} of the power available on this device. For a device like a smart phone, this can be an annoyance, however this level of power consumption could cause catastrophic results in a device such as a pacemaker. In such cases, an alternative is needed for increasing the accuracy without suffering from these short comings. 

Pruning is a tool for reducing the size of a model~\cite{molchanov2019importance,he2017channel}. However, it has only used for this purpose. Although it has been noted in previous works that pruning can result in some minimal increase to accuracy~\cite{bonsai}, the exact circumstances where pruning improves or reduces accuracy has never been understood. Further, previous works have not elucidated how to maximize the desired effect by changing the criteria used to determine which neurons in the network are contributing to the correct predictions. Other methods for reducing the size of a model exist~\cite{Stelzer2021-ci,9927846,Zhu2022-tc}, however they do not add to the accuracy of a model, simply serve as a way to reduce their memory usage.

In this work, we introduce Fine-Pruning, a method that leverages the principles of biological neural pruning to personalize and optimize machine learning models. Inspired by synaptic pruning in the brain, Fine-Pruning utilizes data-driven model compression to selectively remove weights that contribute least to the model's performance on a specific user's data. This approach not only reduces the computational footprint of the model but also enhances its accuracy on the target dataset, all without requiring labeled data or computationally expensive backpropagation. Our research demonstrates the efficacy of Fine-Pruning across a range of applications, including speech recognition and image classification, using popular model architectures such as VGG~\cite{simonyan2015deep}, ResNet\cite{resnet}, and MobileNet~\cite{howard2017mobilenets}. By mimicking the brain's ability to adapt neural architectures based on individual experiences, Fine-Pruning offers a practical solution to the challenges of model personalization and efficiency, particularly for deployment on resource-constrained devices such as smartphones~\cite{apple1} or IoT devices~\cite{8355177}. This biologically inspired approach addresses current limitations in AI deployment and opens new avenues for creating more adaptive and efficient machine learning systems.

\section*{Related Work}

\subsection*{Backpropagation}

Backpropagation~\cite{rumelhart1986} is the primary optimization algorithm used to train neural networks by computing the gradient of a loss function with respect to the weights of the network. Mathematically, the goal is to minimize a loss function \( L \) defined over the network's output predictions \( \hat{y} \) and the true labels \( y \). The network's parameters are updated using some gradient-based optimizers:

\[
\theta_{t+1} = \theta_t - \eta \nabla_\theta L,
\]

where \( \theta \) represents the parameters, \( \eta \) is the learning rate, and \( \nabla_\theta L \) is the gradient of \( L \) with respect to \( \theta \).

Backpropagation computes these gradients by applying the chain rule through each layer of the network. Let \( z_i \) represent the weighted input to a given layer and \( A_i \) the activation at layer \( i \). The forward pass computes:

\[
z_i = W_i A_{i-1} + b_i, \quad A_i = f(z_i),
\]

where \( W_i \) are the weights, \( b_i \) are biases, and \( f \) is the activation function. The loss gradient with respect to the final layer's output is propagated backward through each layer using the chain rule:

\[
\frac{\partial L}{\partial W_i} = \frac{\partial L}{\partial A_i} \frac{\partial A_i}{\partial z_i} \frac{\partial z_i}{\partial W_i}.
\]

Here, \( \frac{\partial L}{\partial A_i} \) depends on the gradient computed from downstream layers, and \( \frac{\partial A_i}{\partial z_i} = f'(z_i) \) represents the activation's derivative. 

The implementation requires storing intermediate activations (computed during the forward pass) to calculate these derivatives efficiently, resulting in significant memory usage, particularly in very deep networks with large batch sizes, making them impractical for working on resource constrained devices.

\subsection*{Pruning}
Neural network pruning removes unnecessary nodes/weights to make the model run more efficiently~\cite{DBLP:journals/corr/LiKDSG16}. Our focus on pruning to achieve a more specialized model for the target data (as opposed to retraining that leverages backpropagation and labeled data), originates from research observations in multi-task learning. A phenomenon that arises is that pruning will cause the model to 'forget' prior features from the source data. This is typically seen as a negative in multi-task learning, however this can be leveraged in favor of personalization.

While we use pruning to increase the accuracy, the main purpose of pruning is to introduce model sparsity, i.e., the percent of elements removed from the original model. There are a few ways this is achieved: either removing blocks of nodes entirely (i.e., structured pruning~\cite{anwar2015structured}), or zeroing out the element and allow for the low level architecture implementation optimize any operations used (i.e., unstructured pruning~\cite{laurent2020revisiting}). This can be done by utilizing sparse matrix libraries like BLAS~\cite{blackford2002updated} on modern IoT and embedded devices. In this paper, we utilize structured pruning based on the aggregated activations of a particular node.

\subsection*{Biological Inspiration.} Our work is inspired by \textbf{biological synaptic pruning}, a phase in the development of the nervous system, is the process of synapse elimination that occurs between early childhood and puberty in many mammals, including humans~\cite{chechik1998synaptic}. During pruning, both the axon (i.e., long projection of a nerve cell - neuron) and dendrite (neuron's extensions that propagate the electrochemical stimulation received from other neurons to the cell body) decay and die off. The infant brain will increase in size by a factor of up to 5 by adulthood, reaching a final size of approximately 86 billion neurons~\cite{alberts2015essential}. The total number of neurons, however, remains the same. After adolescence, the volume of the synaptic connections decreases again due to synaptic pruning specifically for each individual~\cite{craik2006cognition}. Although it is now known that neural synaptic pruning happens for a wide variety of reasons~\cite{Faust2021}, the main mechanism in learning based pruning is activation, which is what our work aims to parallel. 

\section*{Results}

\subsection*{The Fine-Pruning Algorithm}

Fine-Pruning draws inspiration from the biological process of synaptic pruning observed in the developing human brain. This process, which involves the elimination of unused or weak neural connections, counterintuitively enhances cognitive function by reducing noise and focusing neural activity on the most relevant pathways. We apply this principle to artificial neural networks to address the challenge of personalizing models for individual users or specific subsets of data, particularly in resource-constrained environments such as mobile devices. Our method leverages the observation that, just like in the brain, removing redundant or misused weights in an over-parameterized model can increase its accuracy. By treating a general source model as an over-parameterized version of the desired target model, we use pruning as a method to increase accuracy for specific target datasets. This approach involves downloading a generalized model onto various devices and using target data to prune the model, making it both smaller and more accurate (Figure 1).

The core of our method is a data-driven absolute magnitude pruning algorithm. Unlike traditional pruning techniques that rely solely on weight magnitudes, our approach considers the activation of each weight with respect to the target dataset. The algorithm evaluates the source model on the target dataset, records activation maps for each layer, identifies weights that contribute least to the model's predictions on the target data, and prunes these weights to create a sparse sub-model tailored to the target dataset. This process continues until a desired level of sparsity is reached or accuracy begins to decline, mimicking the biological pruning of synapses where the least used neural pathways are eliminated to enhance overall cognitive function. For more information regarding our approach please refer to the Methods section.

\begin{figure*}[ht!]
\centering
\includegraphics[width=.9\textwidth]{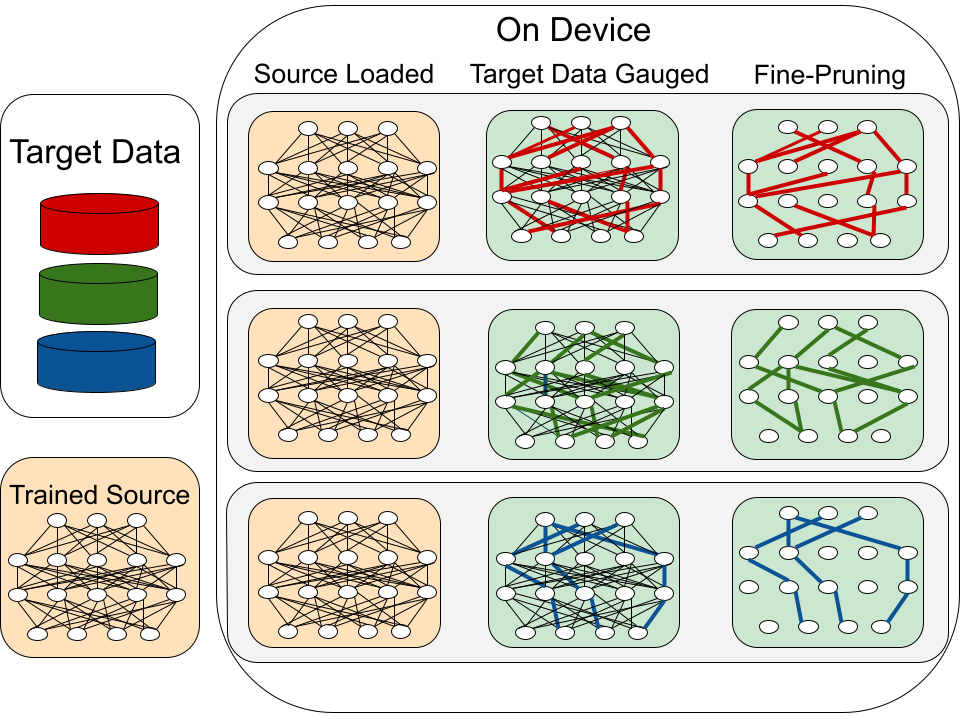}	\caption{The Fine-Pruning algorithm. First, a general selection of data is used to train the model. Second, this generalized model is downloaded onto various devices, each with its own data domain (with no label) which the general model has not seen. The activation energies are recorded. Third, while on board, the model uses the target data to prune itself making the model smaller and more accurate.}
\label{fig:fine_pruning}
\end{figure*}

\subsection*{Comparative Analysis with Existing Methods}

To evaluate Fine-Pruning efficiency and effectiveness compared to existing approaches, we conducted a comprehensive benchmark analysis against both traditional methods and state-of-the-art pruning techniques. We compared Fine-Pruning against standard backpropagation-based fine-tuning, SVD-based model decomposition, and recent pruning methods including DFPC and DepGraph. Our analysis focused on three critical metrics: computational complexity (memory footprint), model compression (induced sparsity), and performance improvement (accuracy gain).

\begin{center}
\begin{tabular}{|c|c|c|c|}
\hline
\textbf{Method} & \textbf{Memory Footprint} & \textbf{Induced Sparsity} & \textbf{Max $\Delta$ Accuracy} \\ 
\hline
\textbf{Backpropagation~\cite{rumelhart1986}} & $O\left( N \cdot \sum_{i=1}^L n_{i-1} \cdot n_i\right)$ &  0\% & 55.2  \\ 
\hline
\textbf{SVD~\cite{6854828}} & $O\left(N \cdot \sum_{i=1}^L \min(n_i n_{i-1}^2, n_i^2 n_{i-1})\right)$ & 0\% & 56.6 \\ 
\hline
\textbf{DFPC~\cite{narshana2023dfpc}} & $O(N \cdot n_{max}^2 + n_{max} \cdot D)*$ & \textbf{81.03\%}  &  -3.38 \\ 
\hline
\textbf{DepGraph~\cite{fang2023depgraphstructuralpruning}} & $O(N \cdot D \cdot n_{max}^3)*$ & 77.39\% & -3.11  \\ 
\hline
\textbf{Fine-Pruning } &  $O\left(\sum_{i=1}^L n_{i-1} \cdot n_i\right)$& 65\% & \textbf{67.2} \\ 
\hline
\end{tabular}
\end{center}

\small{\textbf{Table 1: Comparative Analysis of Model Optimization Methods}. Memory footprint is given in Big-O notation (N = training examples, $L$ = layers, $n_{i}$ = neurons per layer, $n_{max}$ = maximum layer size, $D$ = pruning graph depth, *excluding backpropagation overhead). Results show Fine-Pruning achieves superior accuracy improvement (67.2\%) with moderate sparsity (65\%) while maintaining lower computational complexity. Experiments performed on ResNet50 using ImageNet dataset. The test conditions are the same for starting accuracy, however we are only reporting the individual with the maximum change to accuracy among individuals. Baselines for these values can be found in Figure 2 (a) for initial accuracies used.}

\hfill \break

The results demonstrate several key advantages of Fine-Pruning. First, its memory footprint is notably more efficient than other methods, as it doesn't scale with the number of training examples. This makes Fine-Pruning particularly well-suited for resource-constrained environments. In contrast, traditional methods like backpropagation and SVD require memory proportional to the training set size, while newer pruning approaches like DFPC and DepGraph have even higher computational complexities due to their dependency graph calculations.
Second, while Fine-Pruning achieves a moderate sparsity of 65\%, competing pruning methods DFPC and DepGraph achieve higher sparsity rates (81.03\% and 77.39\% respectively). However, this higher sparsity comes at a significant cost: both methods actually decrease model accuracy on the target dataset (-3.38\% and -3.11\% respectively). This is because these methods are not aimed at increasing accuracy, and prune away important nodes, as will be explained in future results. In contrast, Fine-Pruning demonstrates a remarkable accuracy improvement of 67.2\%, substantially outperforming both traditional methods (backpropagation: 55.2\%, SVD: 56.6\%) and modern pruning approaches.

These results were obtained using ResNet50 on the ImageNet dataset, following the same experimental protocol used in our detailed analysis (Figure 2 (a)). For consistency across methods, we utilized the prescribed protocols and optimizations for each approach, with backpropagation-based fine-tuning specifically using the ADAM optimizer (learning rate = $1 \times 10^{-2}$, batch size = 64, 10 epochs).

\subsection*{Performance Evaluation Across Datasets and Architectures}

To validate the effectiveness of Fine-Pruning, we conducted a series of experiments across various datasets and model architectures, benchmarking our method against existing solutions. Our experiments focused on two main application areas: speech recognition and image classification. For speech recognition, we used the free spoken digit datasets~\cite{FreeSpokenDigitDataset} with VGG-7\cite{wan2018tbn} and MobileNetV2~\cite{howard2017mobilenets} models. For image classification, we employed the CK+ face emotion dataset~\cite{ck+} and ImageNet~\cite{5206848} with VGG-19~\cite{simonyan2015deep} and ResNet50~\cite{resnet} models. We compared Fine-Pruning against two main benchmarks: backpropagation based retraining and SVD-based model decomposition~\cite{LUO2019311}, both of which are common choices for model personalization in resource-constrained environments. The accuracies reported are the target accuracies created by the given model on the give target datasets. 


Results for memory footprint were created utilizing detailed analysis of the algorithms in each work, induced sparsity and Max $\Delta$ Accuracy, the results were taken from experiments done on ResNet50 on the ImageNet dataset with the same procedures for selecting source and target datasets as in Figure 2 (a). As with all tests, we utilize the proscribed protocols and optimizations in the respective works. For fine-tuning utilizing backpropagation, we utilize the ADAM optimizer with a learning rate of $1 \times 10^{-2}$, batch size of 64, and 10 epochs.


Our experiments demonstrate the effectiveness and versatility of Fine-Pruning across various datasets, model architectures, and application domains. We first evaluated Fine-Pruning on a large-scale image classification task using ResNet50 on the ImageNet dataset (Figure 2 (a)). We created "individuals" by selecting pairs of challenging-to-distinguish classes from the ImageNet validation set. Fine-Pruning consistently outperformed the source model, standard backpropagation-based retraining, and SVD-based model decomposition. The improvements were particularly striking for some individuals, where Fine-Pruning achieved near-perfect accuracy from baselines as low as 20-30\%. 

To test Fine-Pruning's performance on a different modality, we applied it to a speech recognition task using a LeNet-5 architecture on the Free-Spoken-Digit dataset (Figure 2 (b)). This dataset contains 500 speech samples per individual, totaling nearly 25 minutes of speech per person. Fine-Pruning again showed superior performance across all individuals tested. The consistent improvement over baseline, backpropagation, and SVD methods highlights the robustness of Fine-Pruning in handling diverse speech patterns and accents. Notably, the method's effectiveness was maintained across individuals with varying baseline accuracies, suggesting its adaptability to different levels of initial model performance.

To further validate the method's versatility, we applied Fine-Pruning to facial emotion recognition using a VGG-19 architecture on the CK+ dataset (Figure 2 (c)). We selected individuals with more than 12 supported emotions, resulting in 14 individuals for our experiments. Here, Fine-Pruning outperformed both the baseline model and standard backpropagation-based personalization. This is particularly significant given the complexity and subtlety of facial emotion expressions, highlighting Fine-Pruning's capability to capture and optimize for fine-grained, user-specific features.

Finally, we investigated Fine-Pruning's performance on a more compact model designed for mobile devices. We used MobileNetV2 on the ImageNet dataset, again utilizing pairs of challenging-to-distinguish classes (Figure 2 (d)). Fine-Pruning again consistently improved accuracy across all individuals, often reaching perfect accuracy even when baseline accuracies were already high (above 95\% for most individuals). It should be noted that SVD did not work for this model architecture due to the interdependencies of the large layers in the model. This highlights another advantage of our method, as it does not suffer from such architectural limitations. These results demonstrate that Fine-Pruning can effectively personalize and improve performance even for compact, mobile-optimized models with initially high accuracies.

\begin{figure}[!ht]
\centering
\begin{subfigure}[b]{0.49\textwidth}
    \centering
    \includegraphics[width=\linewidth]{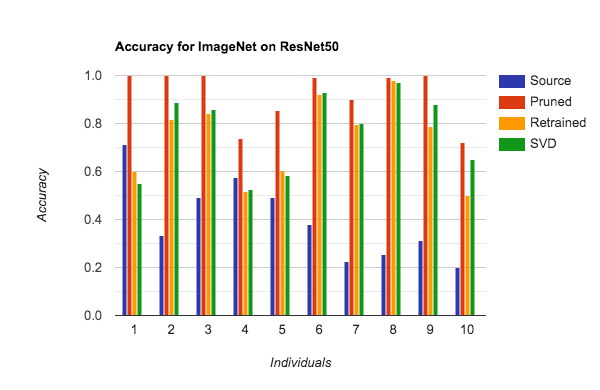}
    \caption{Accuracy for individuals constructed from the ImageNet dataset, tested on ResNet50. }
    \label{fig:resnet}
\end{subfigure}
\hfill
\begin{subfigure}[b]{0.49\textwidth}
    \centering
    \includegraphics[width=\linewidth]{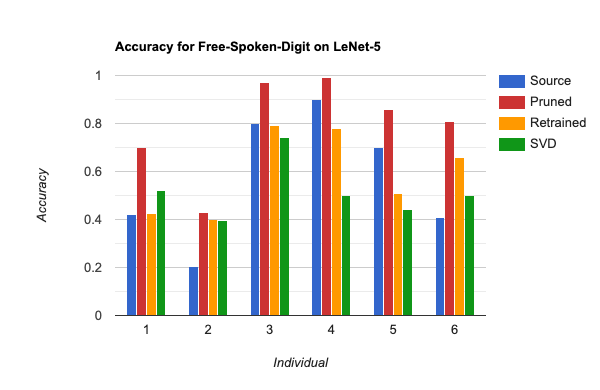}
    \caption{Comparison of individuals' accuracy using the free-spoken-digit dataset on a VGG-7 architecture.}
    \label{fig:acc_vs_ind}
\end{subfigure}

\vspace{0.5cm}

\begin{subfigure}[b]{0.49\textwidth}
    \centering
    \includegraphics[width=\linewidth]{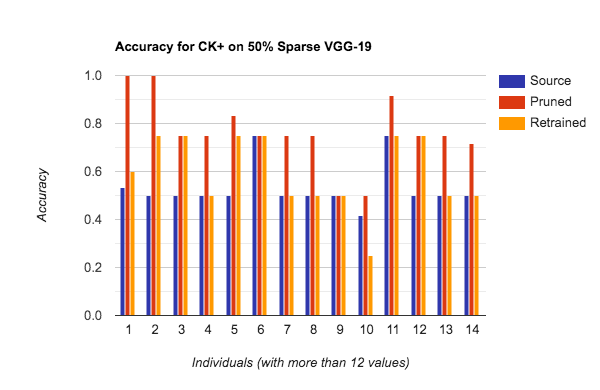}
    \caption{Results on the CK+ facial emotion dataset using a VGG-19 architecture.}
    \label{fig:ck_plus}
\end{subfigure}
\hfill
\begin{subfigure}[b]{0.49\textwidth}
    \centering
    \includegraphics[width=\linewidth]{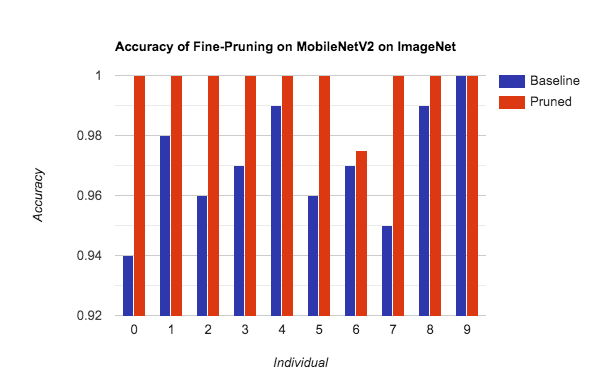}
    \caption{Accuracy of different individuals constructed from ImageNet tested on MobileNetv2 model.}
    \label{fig:mobilenet}
\end{subfigure}
\caption{\textbf{Performance comparison of Fine-Pruning across different datasets and model architectures. 
a.} Accuracy for individuals constructed from the ImageNet dataset, tested on ResNet50. 
Each group represents a different individual, with bars showing baseline accuracy (blue), 
Fine-Pruning (red), standard backpropagation (yellow), and SVD (green). Fine-Pruning 
consistently outperforms other methods, with some individuals seeing dramatic improvements. 
\textbf{b.} Comparison of accuracy across different individuals using the larger free-spoken-digit 
dataset on a VGG-7 architecture. The plot shows baseline accuracy (blue), Fine-Pruning (red), 
backpropagation (yellow), and SVD (green). Fine-Pruning demonstrates consistent improvements 
across all individuals. 
\textbf{c.} Results on the CK+ facial emotion dataset using a VGG-19 architecture. The graph compares 
the baseline model (red), Fine-Pruning (blue), and backpropagation (yellow). Fine-Pruning 
shows superior performance in personalizing the model for individual subjects' facial 
expressions. These results collectively demonstrate the versatility and effectiveness of 
Fine-Pruning across various domains, datasets, and model architectures.
\textbf{d.} Accuracy across different individuals constructed from the ImageNet dataset tested on MobileNetv2 model. Each individual has two classes of the 1000 classes selected from the classes with the highest accuracies in the baseline model.}
\label{fig:performance}
\end{figure}

Our experiments revealed several intriguing characteristics of Fine-Pruning, demonstrating its effectiveness and efficiency across multiple dimensions. We conducted three additional experiments using ResNet50 on the ImageNet dataset to explore the relationship between sparsity and accuracy, data efficiency, and the impact of feature similarity between target and source datasets. First, we investigated how the level of sparsity affects model accuracy (Figure 3 (a)). We gradually increased the sparsity of the model from 0\% to 100\% and measured the resulting accuracy. Interestingly, the target accuracy showed minimal improvement until reaching approximately 65\% sparsity, after which it increased dramatically. Peak performance was achieved at around 70\% pruning. Beyond this point, accuracy declined sharply, reaching the starting accuracy at around 86\% sparsity. This finding suggests that the initial general model was significantly over-parameterized for individual target datasets, despite its strong performance on the broader source dataset. The sharp accuracy decline beyond 70\% sparsity indicates a critical threshold where essential features for the target task begin to be removed. 

Next, we examined the data efficiency of Fine-Pruning compared to conventional retraining with backpropagation (Figure 3 (b)). We varied the percentage of target data used for both Fine-Pruning and backpropagation-based retraining, measuring the resulting accuracy improvements. Fine-Pruning demonstrated superior data efficiency, achieving a 75\% accuracy increase from the baseline using only 60\% of the available target data. In contrast, conventional retraining required over 90\% of the data to reach comparable performance. 

Lastly, we explored how the similarity between target and source dataset distributions affects Fine-Pruning's performance (Figure 3 (c)). We varied the number of classes in the target dataset from 2 to 9 and measured the resulting accuracy. The results indicate that Fine-Pruning's effectiveness is inversely related to the number of classes (features) in the target dataset. As the number of target classes increased, thereby increasing similarity to the source dataset, Fine-Pruning's accuracy showed a slight decline. However, it's important to note that the performance remained high (above 80\%) even with increased feature overlap. This trend suggests that Fine-Pruning is particularly adept at personalizing models to specific subsets of features or classes, which aligns well with many practical applications where end-users typically engage with a limited subset of a model's full capability range.

Collectively, these findings highlight Fine-Pruning's potential as a powerful tool for model personalization, especially in scenarios with limited data availability or where the target task involves a specific subset of the original model's capabilities. The method's ability to achieve high accuracy with minimal data and its effectiveness in creating sparse, task-specific models open new avenues for deploying sophisticated AI systems in resource-constrained environments.

Furthermore, Fine-Pruning offers significant computational advantages. Unlike backpropagation, which requires storing separate activations for each batch, Fine-Pruning only records one activation map that is aggregated after every forward pass. This results in a memory footprint of approximately 1/batchsize that of backpropagation. Additionally, Fine-Pruning uses about half the multiply-accumulate operations (MACs) of backpropagation, as it doesn't require gradient calculations. These efficiency gains, combined with the reduced model size resulting from pruning, make Fine-Pruning particularly well-suited for deployment on resource-constrained devices.

\begin{figure}[!ht]
\centering
\begin{subfigure}[b]{0.32\textwidth}
    \centering
    \includegraphics[width=\linewidth]{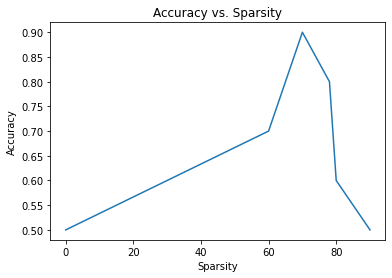}
    \caption{Relationship between model sparsity and accuracy for Fine-Pruning.}
    \label{fig:acc_vs_sparcity}
\end{subfigure}
\hfill
\begin{subfigure}[b]{0.32\textwidth}
    \centering
    \includegraphics[width=\linewidth]{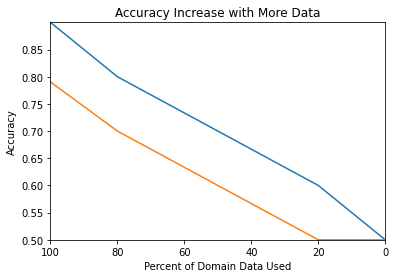}
    \caption{Data efficiency comparison between Fine-Pruning and backpropagation.}
    \label{fig:acc_vs_data}
\end{subfigure}
\hfill
\begin{subfigure}[b]{0.32\textwidth}
    \centering
    \includegraphics[width=\linewidth]{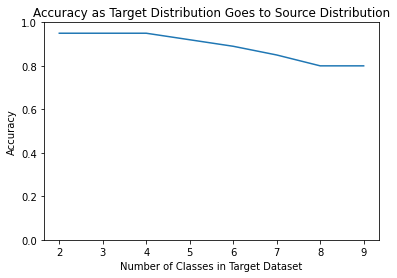}
    \caption{Effect of feature similarity between target and source datasets on Fine-Pruning performance.}
    \label{fig:t_to_s}
\end{subfigure}
\caption{\textbf{Performance characteristics of Fine-Pruning compared to traditional methods.} 
\textbf{a.} The accuracy increases until around 70\% sparsity, after which it begins to decline. 
\textbf{b.} Fine-Pruning (blue) begins increasing accuracy with less data compared to backpropagation (orange). 
\textbf{c.} As the number of target classes increases, the accuracy of Fine-Pruning decreases slightly but remains high. All experiments were performed using ResNet50 on the ImageNet dataset.}
\label{fig:accuracy}
\end{figure}

\section*{Discussion}

Our study introduces Fine-Pruning as a different approach to model personalization, drawing inspiration from the biological process of synaptic pruning. The results demonstrate that this method achieves significant improvements in model accuracy for individual users. In addition, it also addresses key challenges in deploying machine learning models on resource-constrained devices. Here, we discuss the implications of our findings, their relationship to existing work, and potential future directions. The effectiveness of Fine-Pruning across diverse datasets and model architectures suggests a fundamental principle at work: the removal of unnecessary connections can enhance model performance for specific tasks or user profiles. This aligns with observations in neuroscience, where pruning of synaptic connections during development is associated with cognitive refinement. Our finding that optimal performance is achieved at around 70\% sparsity is particularly intriguing, as it mirrors pruning rates observed in biological neural systems~\cite{pmid20972254,Huttenlocher1979-gl,Navlakha2015-jz}. This convergence between artificial and biological systems suggests that there may be universal principles governing efficient information processing in neural networks, whether biological or artificial. Further analysis would be needed to find out the validity of this claim, but it would point towards a greater ability to draw comparisons between live and artificial neural networks and may open new avenues for more biomimicry. 

The ability of Fine-Pruning to improve model accuracy without the need for labeled data or expensive backpropagation represents a significant advancement in the field of personalized machine learning. Traditional methods often require substantial computational resources and large amounts of labeled data, which are frequently unavailable in real-world scenarios, particularly on edge devices. By leveraging only the forward pass information and working with unlabeled data, Fine-Pruning makes on-device personalization feasible for a wide range of applications, from speech recognition to image classification. Our results also shed light on the nature of over-parameterization in neural networks used as source models. The fact that we could achieve substantial sparsity while simultaneously improving accuracy suggests that many trained models contain significant redundancy when applied to specific user data. This redundancy, while potentially beneficial for generalization across a wide range of users, can be detrimental when targeting individual use cases. Fine-Pruning effectively navigates this trade-off, retaining the benefits of a well-trained general model while optimizing for individual performance. The superior data efficiency of Fine-Pruning compared to conventional retraining methods is particularly noteworthy. In many real-world applications, the amount of data available for personalization is limited. Our method's ability to achieve significant improvements with less data makes it more practical for deployment, and also, potentially, more privacy-preserving, as it requires less user data to be collected or processed, and further it does not require for the data or model to be transmitted to another device for retraining~\cite{yao2007predicting}. The relationship we observed between feature sparsity in the target dataset and the effectiveness of Fine-Pruning provides valuable insights for potential applications. The method appears to be especially powerful when personalizing models for users or tasks that utilize a subset of the full feature space of the original model. This characteristic makes Fine-Pruning particularly well-suited for applications where users typically engage with a limited portion of a model's capabilities, such as personalized speech recognition or specialized image classification tasks. From a computational perspective, the reduction in model size and complexity achieved through Fine-Pruning has significant implications for deploying AI on edge devices. The decreased memory footprint and computational requirements make it feasible to run sophisticated models on resource-constrained devices and contribute to energy efficiency, a critical factor in mobile and IoT applications~\cite{husak2018survey}. This aligns with the growing trend towards edge AI, where processing is done locally on devices rather than in the cloud, offering benefits in terms of latency, privacy, and offline functionality.

Our comparative analysis against both traditional methods and state-of-the-art pruning techniques further validates Fine-Pruning's practical advantages. While newer pruning methods like DFPC and DepGraph achieve higher sparsity rates (\>77\%), they do so at the cost of accuracy, showing performance degradation on target datasets. This trade-off between sparsity and accuracy represents a significant challenge in existing pruning approaches. Fine-Pruning, in contrast, demonstrates that moderate sparsity (65\%) can be achieved while simultaneously improving accuracy by over 67\% – substantially outperforming both traditional methods like backpropagation and SVD, which show accuracy improvements around 55\%. Perhaps more significantly, Fine-Pruning achieves this with markedly lower computational complexity, as it doesn't scale with the number of training examples. This combination of improved accuracy, reasonable sparsity, and computational efficiency makes Fine-Pruning particularly well-suited for real-world applications, especially in resource-constrained environments like mobile and IoT devices where both model performance and computational efficiency are critical.

While our results are promising, they also open up new questions and avenues for future research. One key area of investigation is the generalizability of Fine-Pruning across even more diverse types of neural network architectures and tasks. While we've demonstrated its effectiveness in CNNs for speech and image tasks, exploring its applicability to other architectures like transformers or recurrent neural networks could further broaden its impact. Another important direction is related to the theoretical foundations of why Fine-Pruning works so well. Understanding the mathematical principles underlying the relationship between pruning, personalization, and accuracy could lead to even more effective algorithms and potentially inform the design of new neural network architectures that are inherently more amenable to personalization. The potential of Fine-Pruning in federated learning scenarios is also an exciting area for future work. Given its ability to create personalized models without sharing raw data, Fine-Pruning could be integrated into federated learning systems to create more efficient and privacy-preserving distributed learning frameworks. Lastly, the biological inspiration behind Fine-Pruning suggests that further exploration of neuroscience-inspired algorithms could yield additional breakthroughs in AI. The success of this approach invites us to consider what other principles from biological learning systems might be advantageously applied to artificial neural networks.

In conclusion, Fine-Pruning represents a significant step forward in creating more efficient, accurate, and personalized AI models, particularly for resource-constrained environments. By bridging insights from neuroscience with advanced machine learning techniques, our work offers a practical solution to current challenges in AI deployment and opens up new perspectives on the nature of learning and adaptation in neural systems. As we continue to push the boundaries of AI, approaches like Fine-Pruning that draw inspiration from biological systems may play an increasingly crucial role in developing more adaptive, efficient, and personalized machine learning technologies.

\subsection*{Limitation}

Fine-pruning cannot train a model from nothing or randomized weight. While it is possible that this method may be able to extract some accuracy from a model that has not been trained on any data whatsoever, this falls outside the scope of this work. Fine-pruning is designed only to work on personalizing, or fine tuning, a large source model to perform better on a target dataset. While this may seem like a strong precondition, we see this in a multitude of real world applications, such as a general speech model that is being personalized on an individual user's device for their specific voice. 

As can be seen in Figure 3 (c), as the number of classes grows the peak accuracy tends toward the starting accuracy of the model. In other words, as the number of features in the target dataset approaches the number of features in the source dataset, the benefits to accuracy of Fine-Pruning decrease. We are not aware of any parallel to this phenomenon from our biological background. That being said, even if the accuracy doesn't improve, we still see the other benefits of conventional pruning. Those being the reduction of memory usage and energy consumption. The exact point where the accuracy increase is no longer seen as the target distribution's features increases has not been observed to be a universal constant across datasets, however from our experiments the amount seems to be far less than those seen in a multitude of different real world domains, as shown in our results.

Further, as seen explained in our analysis of Figure 3 (a), after the optimal pruning rate is found, the decrease to accuracy is sharp. This means that if you overshoot the optimal threshold you may find the resulting accuracy is less optimal, or perhaps even below the starting accuracy. While our study, as well as the biological inspiration, seems to suggest that a pruning rate from 60\% - 70\% works broadly for different models and datasets, it may perhaps be safer to start with a lower amount and build up until there is observed decrease to accuracy. Since the distribution is mono-modal, and the cost for pruning is very small once the activation maps are tabulated, this should not be a costly procedure.

\section*{Methods}

\subsection*{Fine-Pruning Algorithm}

This work uses the following definitions:
\begin{itemize}
    \item \noindent\textbf{Source data} is dataset for training that the model initially is learned over. These data points are, for instance, speech samples from a large number of individuals, but not the individual (e.g., smartphone owner) that will ultimately use the model.

    \item \noindent\textbf{Target data} is what the model has to specialize to on each device. These data are later produced by the end-users on each device (e.g., speech samples of a given phone owner) and are not labeled. 
    
    \item \noindent\textbf{Target accuracy} refers to the model accuracy on the target data. This will differ from 'accuracy', which will be withheld for accuracy over the entire dataset (i.e. both source and target data).

\end{itemize}

Our study introduces Fine-Pruning, an alternative approach to personalizing trained neural network models. This method draws inspiration from biological synaptic pruning, aiming to optimize model performance for specific target datasets while reducing computational complexity.

The core of our approach is the Fine-Pruning algorithm, which leverages data-driven absolute magnitude pruning. Given a trained source model $N_{Source}$ and a target dataset $T$, we first evaluate $N_{Source}$ on $T$. For each layer in $N_{Source}$, we compute an activation map representing the contribution of each weight to the model's output for the target dataset. We then define a pruning threshold $\sigma$, which determines the proportion of weights to be removed.

The pruning operation can be formally defined as:

\begin{equation}
prune(\sigma, N_{Source}, T) = [
\begin{cases}
\hat{w_i} = 0 & \text{if } \sum_{t \in T} |A(w_i)| < \sigma \text{ and } |N_{Source}| > T \\
\hat{w_i} = w_i & \text{otherwise}
\end{cases}
: \forall w_i \in N_{Source} ]
\end{equation}

%


where activation map $A$ consists of elements $A(w_i)$, calculated by summing the activation of weight $w_i$ in source model $N_{Source}$ across all target input values $t$ in dataset $T$, and $\hat{w_i}$ is the weight in the pruned model.

If the source model has the ability to learn a wide variety of diverse features, specifically enough to capture the particular features of the individuals with their target data, then the weights utilized to discern between features which are not present in a target data become redundant for that individual. These unused weights contribute nothing to the target model's performance, and in fact create noise that hinders the accuracy. Note, we require that the model be trained on the source data to begin with, as the brain has some initial connections that are used before being pruned away. 

In more computer science terms, our proposed Fine-pruning is similar in concept to the lottery ticket hypothesis~\cite{lottery}. If we view the target network as the \textit{winning ticket} within the source model, then this work's goal can be represented in a like fashion to the work done in the lottery ticket hypothesis. The difference between this work and the lottery ticket hypothesis is that this work does not look to remain accurate on the source data, but rather optimize and personalize for the target data.

More specifically, if we have a data domain of $X$, our objective is to search within an over-parameterized source network $N_{Source}$, trained on a source dataset $S \subset X$, for a sub-network $N_{Target} \subset N_{Source}$, which is tailored towards some target dataset $T \subset X$. In order to find $N_{Target}$, we evaluate $N_{Source}[T]$ and consider the activation map of each layer $l$ of the network, $A_l$.

For each $A_l$, we select the indices $i$ that contribute the least to the prediction of $N_{Source}$ over the target dataset.

These indices $i$ are just contributing noise to the prediction, hence removing the nodes associated with $i$ in $N_{Source}$ will reduce the noise with regards to $T$. This will increase the accuracy until the sparsity gets to the point, where further pruning will remove the essential weights that are needed to predict $T$.  

Since $N_{Target} \subset N_{Source}$, the solution prunes away all unnecessary weights until we arrive at  
\begin{equation}N_{Target} = \min_{\sigma}prune(\sigma,N_{Source},T).
\end{equation}
Even with a $\sigma$ pruning threshold that is not the optimal value, $prune(\sigma,N_{Source},T)$ will be an approximation for $N_{Target}$, and increases the the target accuracy. 

The pseudo-code for the Fine-Pruning algorithm is as follows:
\begin{algorithm}
\caption{Fine-Pruning algorithm }
\label{alg:fine-pruning}
\begin{flushleft}
\hspace{0cm} \textbf{Input:} Model $N$, Dataset $D$, Float $sp$ \\
\hspace{0cm} \textbf{Output:}  Model $N$ \\
\hspace{0cm} ActivationMap $A = $ \{\}  \\
\hspace{0cm} $\forall$ Sample $s \in D$: \\
\hspace{.5cm} Evaluate $N$ on $s$ \\
\hspace{.5cm} $\forall$ Layers $l \in$ evaluation: \\
\hspace{1cm} $A$[index of $l$] += $l$ \\
\hspace{0cm} $\forall$ Layers $l \in A$: \\
\hspace{.5cm}  $\forall$ Weights $w \in l$: \\
\hspace{1cm} if \% weights $>$ $sp$ \&  $min(A) = w$: \\
\hspace{1cm} prune $w$ in $N$ 
\end{flushleft}
\end{algorithm}

\subsection*{Datasets and Preprocessing}

To evaluate the effectiveness of Fine-Pruning, we conducted experiments across multiple datasets in two domains: speech recognition and image classification. 

For speech recognition, we used the Free Spoken Digit Dataset~\cite{FreeSpokenDigitDataset}. The Free Spoken Digit Dataset is larger, containing 500 speech samples per individual, totaling nearly 25 minutes of speech per person. We preprocessed these datasets by converting audio samples into 2D spectrograms, enabling the use of convolutional neural networks.

For image classification, we employed the CK+ dataset for facial emotion recognition and a subset of ImageNet for general object classification. The CK+ dataset contains facial emotion images, from which we selected individuals with more than 12 supported emotions, resulting in 14 individuals for our experiments. For ImageNet, we created "individuals" by selecting two random classes from the validation set for each personalization experiment.

\subsection*{Model Architectures and Experimental Setup}

We tested Fine-Pruning on several popular neural network architectures: VGG-7 for speech recognition tasks, VGG-19 for the CK+ facial emotion recognition task, ResNet50 for ImageNet classification, and MobileNetV2 to demonstrate Fine-Pruning's effectiveness on mobile-optimized architectures. All models were implemented using Keras with a TensorFlow backend, and trained weights were used as starting points where applicable.

Our experimental procedure involved training or obtaining a trained model on the full source dataset, selecting a subset of the data to serve as the target dataset (simulating an individual user's data), applying Fine-Pruning to the trained model using only the target dataset, and evaluating the pruned model's performance on a held-out portion of the target dataset. We used 5-fold cross-validation for all experiments to ensure robust results.

\subsection*{Benchmarking and Evaluation}

To benchmark Fine-Pruning's performance, we compared it against two main alternatives: backpropagation based retraining and SVD-based model decomposition. For backpropagation based retraining, we fine-tuned the trained models on the target datasets using standard backpropagation with the Adam optimizer. The SVD-based method followed the approach described in previous work on model complexity reduction for personalization.

We evaluated performance using several metrics: accuracy (the primary metric for all classification tasks), sparsity (the percentage of weights set to zero in the pruned model), data efficiency (the amount of target data required to achieve a given level of accuracy improvement), and computational efficiency (measured in terms of FLOPs required for both the pruning process and inference with the pruned model).

\subsection*{Implementation Details and Statistical Analysis}

All experiments were conducted on a system with NVIDIA Tesla V100 GPUs, using Python 3.7, TensorFlow 2.4, and Keras 2.4. The TensorFlow Model Optimization library was used for pruning operations. To ensure statistical significance, we performed paired t-tests comparing Fine-Pruning against each benchmark method for each experiment, considering results significant at p < 0.05.

The datasets used in this study are publicly available, and the code used to implement Fine-Pruning and reproduce our experiments is available at our project repository. Through this comprehensive methodology, we aimed to thoroughly evaluate the effectiveness and efficiency of Fine-Pruning across a wide range of applications and model architectures.

\section*{Broader Impact}

The success of Fine-Pruning demonstrates how biomimicy can guide practical solutions in machine learning, extending beyond the basic neural inspiration of ANNs to incorporate established neurological processes. While the primary contribution of this work is in model personalization, which has the capacity to revolutionize how personalizing devices is done for mobile or IoT, our analysis suggests additional potential applications that merit further investigation.

First, Fine-Pruning may contribute to privacy-preserving federated learning. In distributed scenarios where a general model is deployed across multiple devices (e.g., smartphones), users typically must share either their private data or model gradients to improve the general model's accuracy. Since Fine-Pruning creates personalized models that are strict subsets of the original model's weights, it enables an alternative update mechanism: devices can communicate their pruning masks rather than raw data or full gradient information. This approach could reduce communication overhead while improving user privacy.

Second, the properties of Fine-Pruning suggest potential applications in user authentication through model fingerprinting. The pruning patterns created for each user's specific data distribution generate distinctive masks that could serve as identifiers. Unlike hardware-based keys tied to specific devices, pruning-based signatures would be reproducible across devices while remaining specific to each user's data characteristics. This mechanism warrants investigation for applications in distributed authentication systems.

These potential applications, while requiring  validation, indicate that Fine-Pruning impact may extend beyond its core function of model personalization. More broadly, this work provides evidence that examining biological systems can yield practical solutions for contemporary challenges in machine learning, particularly in areas such as model efficiency and adaptability.

The success of this algorithm serves as a reminder that there are many biological parallels that can inspire solutions for practical, real-world problems in machine learning. Although ANNs are loosely based off of real neurons in the brain, this work shows there are plenty of lessons to learned from real neurological processes. We further propose that Fine-pruning may provide more benefits outside of fine-tuning or pruning, which has the capacity to revolutionize how personalizing devices is done for mobile or IoT, that could merit further investigation. 

For example, end-user privacy and communication efficiency is desired in the use-case of distributing a general model to different target devices (e.g., smartphones), where individual users have their own private dataset (e.g., speech) as the new knowledge to improve the general model’s accuracy. However, sending the entire target dataset is costly and could violate user privacy. There are methods to send only the local gradients to the general model, but this could still be costly in practice. Since each remaining target model is a sub-model of the general model (i.e., Fine-pruning does not create any new weights), any retraining that occurs after the model is pruned can be utilized to enhance the accuracy of the general model. 

Trustworthy end-user fingerprinting is another possible benefit of Fine-pruning for model personalization. With user-specific pruning masks, Fine-pruning essentially creates a unique, reproducible keys (user fingerprints) that can be recreated on different devices. If the models of any user is sufficiently different from all others, then the mask of the unpruned weights can be utilized like a fingerprint for the user. This is potentially better than a hardware key, which is tied to a device, whereas this method generates the same key given the same target data domain or same features that need to be extracted for each user.

These possible continuations of this work would have wide reaching implications in the space of computer engineering, reaching further than just the field of biomimicy.

\section*{Conclusion}
In this paper, we introduce Fine-Pruning, a biologically inspired algorithm that emulates the neural pruning observed in biological systems, where connections are selectively pruned to enhance efficiency and accuracy. Fine-Pruning utilizes a method inspired by absolute magnitude pruning, akin to how biological neural networks adapt and specialize by fine-tuning based on observed stimuli, rather than extensive retraining. This approach fine-tunes an already trained model directly on local end-devices without requiring annotated data labels, thus mirroring the biological principle of adaptation without a known target outcome. Unlike traditional methods that rely on computationally intensive backpropagation, Fine-Pruning operates using only lightweight forward pass information, significantly reducing both computational and memory demands during model fine-tuning and subsequent target inference. Demonstrating its effectiveness, Fine-Pruning enhances the accuracy of a different deep learning model from 50\% to nearly 90\% on local datasets, all while achieving over 70\% model sparsity with minimal computational overhead and no need for data labels. This highlights Fine-Pruning’s potential to advance personalized machine learning by incorporating biologically inspired, resource-efficient algorithms.



\section*{Resource availability}


\subsubsection*{Lead contact}


Requests for further information and resources should be directed to and will be fulfilled by the lead contact, Joseph Bingham (jbingham@technion.ac.il).

\subsubsection*{Data and code availability}


The datasets used in this study are publicly available: \\
- Free Spoken Digit Dataset: \url{https://github.com/Jakobovski/free-spoken-digit-dataset}~\cite{FreeSpokenDigitDataset} \\
- CK+ Dataset: \url{https://paperswithcode.com/dataset/ck}~\cite{ck+} \\
- ImageNet: \url{https://www.image-net.org/}~\cite{ILSVRC15}

The code used to implement Fine-Pruning and reproduce our experiments is available at \url{https://github.com/JosephBingham/fine_pruning_ck-}
with DOI: 10.5281/zenodo.14886693 at Zenodo~\cite{code}.

\section*{Acknowledgments}


 JB and SZ was funded by The National Science Foundation (NSF) and the Department of Energy (DOE) - Office of Cybersecurity, Energy Security, and Emergency Response (CESER) Award DE-CR0000056.
Author DA is supported by the Azrieli Foundation and supported by grant from the Israel Science Foundation (1543/21).
\section*{Author contributions}


JB and SZ initiated the study. JB developed the method, performed the analyses, and drafted the manuscript. DA worked on the final version of the manuscript. All authors reviewed and approved the final manuscript. 

\section*{Declaration of interests}


 DA reports consulting fees from Carelon Digital Platforms and Link Cell Therapies. 

\section*{Declaration of generative AI and AI-assisted technologies}


None of the writing of this work was generated with AI. Formatting of figures in latex was assisted utilizing ChatGPT. 

\section*{Legend}
\subsection*{Figures}
\begin{itemize}
    \item Figure 1: The Fine-Pruning algorithm. First, a general selection of data is used to train the model. Second, this generalized model is downloaded onto various devices, each with its own data domain (with no label) which the general model has not seen. The activation energies are recorded. Third, while on board, the model uses the target data to prune itself making the model smaller and more accurate.
    \item Figure 2: Performance comparison of Fine-Pruning across different datasets and model architectures.
These results collectively demonstrate the versatility and effectiveness of Fine-Pruning across various domains, datasets, and model architectures. (a) Accuracy for individuals constructed from the ImageNet dataset, tested on ResNet50. (b) Comparison of individuals' accuracy using the free-spoken-digit dataset on a VGG-7 architecture. (c) Results on the CK+ facial emotion dataset using a VGG-19 architecture. (d) Accuracy of different individuals constructed from ImageNet tested on MobileNetv2 model.

    \item Figure 3: Performance characteristics of Fine-Pruning compared to traditional methods.
(a) The accuracy increases until around 70\% sparsity, after which it begins to decline. 
(b) Fine-Pruning (blue) begins increasing accuracy with less data compared to backpropagation (orange). 
(c) As the number of target classes increases, the accuracy of Fine-Pruning decreases slightly but remains high. All experiments were performed using ResNet50 on the ImageNet dataset.
\end{itemize}

\subsection*{Tables}
\begin{itemize}
    \item Table 1: Comparative Analysis of Model Optimization Methods.
\end{itemize}

\subsection*{Equations}
\begin{itemize}
    \item Equation 1: The set-theoretical representation of Fine-Pruning
    \item Equation 2: Our proposed definition of finding the best model using Fine-Pruning
    \item Note, there are many other non-numbered equations in this work that are taken from other resources in order to explain other methods. They do not come from us and are only utilized within the context they are stated. As such instead of trying to caption each one of them, we refer you back to the citations provided.
\end{itemize}

\subsection*{Algorithms}
\begin{itemize}
    \item Algorithm 1: Fine-Pruning algorithm
\end{itemize}

\newpage


\bibliography{references}

\end{document}